# Benchmarking Frontier Large Language Models Against Official Crash Database Coding Using Police Crash Narratives

**Sudhir Bharati[1]*, Rajendra K. C. Khatri[2], Sudip Bharati[3]**

[1] Planning and Research Division, Arkansas Department of Transportation, Little Rock, Arkansas, USA
[2] Department of Mathematics, Philander Smith University, Little Rock, Arkansas, USA
[3] Itahari International College, Itahari, Nepal
**Corresponding author: sudhir.bharati@ardot.gov*

## ABSTRACT

Police crash narratives contain information that may supplement structured crash databases, but manual review is labor-intensive and it remains unclear how well large language models (LLMs) reproduce official crash coding. This study benchmarked six frontier LLMs by comparing narrative-derived crash attribute codes with corresponding fields in the Arkansas fatal-crash database. The analysis linked 5,587 fatal-crash narratives with 5,889 structured crash records from Arkansas (2015-2025), yielding 4,194 matched crashes. Six LLMs were evaluated using an identical zero-shot prompt to code crash manner, non-motorist relation, intersection type, work-zone relation, roadway surface condition, and light condition. Performance was evaluated using agreement, macro-averaged F1 score, Cohen's kappa, coverage, selective agreement, and comparisons with always-majority, always-Unknown, and keyword-rule baselines. Repeated-measures analyses and a generalized estimating equations model assessed differences among models and attributes. GPT-5.5 High achieved the highest agreement among the evaluated LLMs, but the always-majority baseline produced higher raw agreement and the keyword-rule baseline achieved macro-averaged F1 score and Cohen's kappa comparable to the best-performing LLM. Agreement was highest for non-motorist relation and crash manner and lowest for light condition, roadway surface condition, and work-zone relation. Differences across crash attributes exceeded differences across models. These results provide a benchmark for evaluating LLM-based crash coding and show that deployment should be evaluated on an attribute-specific basis using transparent baselines and human review.

**Keywords:** *Large language models; police crash narratives; crash databases; transportation safety; natural language processing; benchmarking.*

## INTRODUCTION

Crash databases support much of transportation safety research and practice. Researchers use crash records to identify risk factors, model crash severity, evaluate safety treatments, and investigate the conditions associated with severe outcomes. Transportation agencies rely on the same data to prioritize investments, allocate resources, design countermeasures, and support roadway safety policies. Most of these applications depend on structured attributes that describe crash type, contributing factors, driver actions, roadway characteristics, environmental conditions, and vehicle movements. When these attributes are incomplete, inconsistent, or incorrectly coded, safety analyses may misidentify risk factors, underestimate important crash patterns, or support ineffective countermeasures.

Prior research has shown that official crash databases can be affected by under-reporting, missing information, inconsistent definitions, and coding variability across jurisdictions and reporting systems [1], [2], [3], [4], [5]. These data quality concerns are particularly important because structured crash variables often serve as the foundation for predictive modeling, hotspot identification, trend analysis, and program evaluation. Reliable crash coding is not simply a database management concern. It directly affects the validity of safety analyses and the decisions based on them.

Police crash narratives are another source of crash information. Investigating officers use sentences to describe how a crash occurred and the circumstances surrounding it. Structured fields reduce these circumstances to predefined codes, but narratives may retain details about the sequence of events, vehicle movements, driver behavior, roadway conditions, environmental factors, and interactions among road users. These narrative details are particularly valuable for vulnerable road users, including pedestrians and cyclists, because studies evaluating bicycle infrastructure frequently rely on crash characteristics such as non-motorist involvement, crash manner, and intersection context to assess roadway safety and identify design-related crash patterns [6]. Prior studies have shown that crash narratives can contain information that is missing, simplified, or misclassified in structured fields and can support the identification of crash types, contributing factors, and safety-relevant patterns [7], [8], [9], [10], [11]. Manual review, however, requires substantial time and may produce inconsistent results across reviewers.

Automated text analysis can reduce the effort required to review large narrative datasets. Transportation researchers have applied natural language processing, machine learning, transformer models, and semantic search methods to extract or classify information from crash narratives [7], [8], [9], [10], [11]. Large language models (LLMs) extend these methods by interpreting context and producing structured responses from unstructured text. Recent transportation studies have examined LLMs for safety analysis, crash prediction, information extraction, and decision support [12], [13], [14].

The ability to generate a structured response does not establish that the response will align with official crash coding. Most crash narrative studies have focused on selected crash types, individual variables, or narrowly defined classification tasks [15], [16]. Consequently, the literature has not clearly established which official crash fields show consistently high or low concordance with classifications derived from narrative text. Agreement may vary because some attributes are described more directly in narratives than others, although the present study does not independently assess whether sufficient evidence was available for each coding decision. A disagreement between LLM-generated coding and an official database may therefore reflect model error, error in the official record, record-linkage uncertainty, differences between the information represented in the narrative and structured fields, or insufficient narrative detail.

Generative LLMs also raise reliability concerns that extend beyond conventional classification accuracy. Unlike traditional classifiers that consistently assign observations to predefined categories, LLMs may either produce a specific classification or abstain when the available text is insufficient or ambiguous. Prior transportation LLM studies have identified hallucination, uncertainty, robustness, and trustworthiness as important concerns for safety-critical applications [12], [17], [18], [19]. In crash narrative coding, a key operational issue is therefore how often models assign a specific code versus return an Unknown value. Because an Unknown response may reflect appropriate caution, excessive abstention, or differences in model-specific decision thresholds, the present study evaluates abstention behavior and agreement among committed outputs without interpreting abstention as direct evidence of uncertainty recognition.

This study evaluates large language models (LLMs) as structured crash attribute coders using Arkansas fatal-crash narratives from 2015 through 2025. Six frontier LLMs were applied to a common zero-shot extraction task, and their coded outputs were compared with the corresponding fields in the official structured crash database. The study is framed as an agreement benchmark: it evaluates how closely model-generated coding matches official coding and how models behave when they do not assign a specific code.

The study makes four contributions. First, it benchmarks multiple contemporary LLMs on a common crash attribute coding task using a large matched dataset of Arkansas fatal-crash narratives and structured records. Second, it measures agreement with official coding for each evaluated crash attribute, showing where performance is relatively high and where it is consistently low. Third, it compares model performance with transparent baselines and examines differences across attributes as well as across models. Fourth, it evaluates abstention behavior through coverage and selective agreement, providing a clearer picture of how often models commit to a code and how often those committed outputs agree with the official database. Together, these results clarify the current capabilities and limitations of LLM-assisted crash coding for transportation safety applications.

The study is guided by three research questions. First, how closely do large language models agree with official crash database coding when extracting structured attributes from fatal crash narratives? Second, how does agreement with official coding vary across crash attributes, and are differences across attributes larger than differences across models? Third, how do the evaluated systems differ in coverage, abstention, and agreement among non-abstained outputs?

## LITERATURE REVIEW

Crash data are the empirical foundation of roadway safety research, but the literature has long shown that official crash databases are imperfect representations of crash events. Police-reported data can suffer from under-reporting, reporting bias, missing information, inconsistent coding, and differences in how crashes and casualties are defined across jurisdictions [1], [2], [3], [4], [5], [20], [21]. These limitations are not merely administrative. They influence the apparent distribution of crash types, injury severity, road-user involvement, and contributing factors, and therefore shape the conclusions drawn from safety studies and countermeasure evaluations. Studies linking police records with medical, hospital, emergency room, and registry data have repeatedly shown that official crash datasets may miss substantial portions of crashes or casualties, particularly non-fatal crashes, vulnerable road users, single-vehicle events, and cases with lower injury severity [2], [3], [5]. More broadly, crash data quality reviews have emphasized that roadway safety research depends on accurate, complete, and well-integrated data, yet many databases remain affected by missing values, inconsistent definitions, limited linkage, and unreliable contributory factor coding [4], [18], [20].

Because structured crash fields may be incomplete or inconsistently coded, researchers have increasingly turned to crash narratives as an alternative source of safety information. Narratives are written in natural language and often contain details that structured fields omit, simplify, or misclassify. Early work in natural language understanding showed that police accident descriptions could be transformed into structured representations and compared with coded crash records. Wu and Heydecker demonstrated that plain-English road accident descriptions could be used to recover information about crash location, vehicle movement, impact position, and contradictions between coded and narrative-derived information [10]. This early work established an important idea that continues to motivate current research: crash narratives are not just explanatory notes attached to a record; they are a potentially recoverable source of structured crash information.

Subsequent studies expanded this idea by using text mining, machine learning, and natural language processing to identify crash attributes or safety-relevant patterns that are difficult to capture through structured fields alone. Work-zone crash research showed that narratives could identify misclassified work-zone crashes that were missed by the official work-zone indicator [9]. Wrong-way driving research used police narratives to distinguish actual wrong-way driving crashes from candidate cases, with Bidirectional Encoder Representations from Transformers (BERT) outperforming conventional classifiers [8]. Secondary crash research similarly showed that narratives could improve identification of secondary crashes when structured fields were insufficient or unreliable [11]. Agricultural crash studies demonstrated that manually labeled crash narratives and machine learning models could identify farm-equipment-related crashes across states and future-year data [22]. Other studies used narratives to classify injury severity, highway-rail crash patterns, motorcycle crash causation themes, and broader crash factors using methods such as Term Frequency-Inverse Document Frequency (TF-IDF), Word2Vec, topic modeling, interpretable machine learning, and Random Forests [16], [23], [24], [25]. Together, these studies show that crash narratives can recover meaningful safety information that is either absent from or only weakly represented in structured databases.

However, the literature has often approached narrative text through narrow extraction or classification tasks. Many studies focus on one target attribute, such as work-zone involvement, wrong-way driving, secondary crash status, agricultural crash involvement, injury severity, motorcycle causation themes, or highway-rail crash type [8], [9], [11], [16], [22], [24], [25]. Other studies use narratives to discover latent topics, recurring crash themes, or severity-associated words rather than to reproduce official crash coding variable by variable [15], [16], [23], [26], [27], [28]. These studies are valuable because they demonstrate that narratives contain safety-relevant information. Yet they do not fully answer a different and more operational question: which official crash attributes can actually be recovered from narrative text, and which cannot? This distinction is central for transportation safety research because a narrative may support some coding decisions, partially support others, and provide no evidence for many variables that appear in structured crash databases.

The methodological trajectory of this literature has moved from rule-based systems and keyword extraction toward machine learning, transformer models, semantic search, and more recently large language models. Earlier work relied on knowledge-based parsing, expert-defined rules, domain-specific dictionaries, keyword matching, n-grams, TF-IDF, and traditional classifiers [9], [10], [11], [22], [25], [29]. These methods can work well when the target concept has consistent language, but they are vulnerable to spelling variation, short narratives, ambiguous phrasing, state-specific reporting styles, negation, and context-dependent meanings [7], [9], [10], [11], [22], [25]. More recent studies have used transformer-based models and semantic representations to reduce reliance on exact keyword overlap. BERT improved wrong-way driving classification [8], and semantic search using Sentence Transformer embeddings showed that relevant crash narratives could be retrieved even when exact query terms were absent [7]. At the same time, semantic search introduced its own challenges, including sensitivity to query wording, ambiguity, negation, rare domain-specific expressions, and lexical overlap without intended meaning [7].

This evolution has opened the door to large language models, which are increasingly being considered for transportation safety and mobility applications. Recent reviews describe LLMs as promising tools for extracting, summarizing, reasoning over, and converting unstructured transportation data into structured formats [12], [19]. Customized LLM frameworks such as TrafficSafe and SafeTraffic Copilot have also been proposed for event-level crash prediction and safety intervention analysis using textualized multimodal crash-event information, with prediction targets such as injury count, crash severity, and crash type [14], [30]. More directly related to crash narratives, recent work using National Highway Traffic Safety Administration (NHTSA) Crash Investigation Sampling System narratives evaluated open-source LLMs and fine-tuned models for extracting crash reconstruction variables such as vehicle identifiers, initial travel direction, maneuver, and critical pre-crash event [13], [13]. These studies move the field closer to structured crash narrative coding and show that LLMs can support information extraction from complex crash descriptions.

Yet the use of LLMs also changes the reliability problem. Traditional text classifiers generally make bounded predictions over predefined labels. LLMs, by contrast, may generate structured answers, explanations, or classifications that appear plausible even when the narrative does not support them. This raises concerns about uncertainty recognition, unsupported inference, and hallucination. Transportation LLM reviews identify hallucination, interpretability, privacy, computational cost, and robustness as major barriers to safety-critical deployment [12], [19]. LLM benchmarking studies using real public transportation incident records show that hallucinations can emerge in spatial and temporal reasoning tasks, including plausible but incorrect locations, time calculations, durations, event counts, and unsupported extrapolations beyond the evidence provided [17]. These findings are important for crash narrative extraction because many crash variables require careful distinction between what is explicitly stated, what can be reasonably inferred, and what is not recoverable from the narrative.

Reliability has been a recurring concern across the broader narrative-mining literature, although it has usually been measured as model performance rather than as evidence-supported coding. Prior studies have used manual review, expert labeling, cross-validation, held-out datasets, future-year validation, cross-state validation, confusion matrices, precision, recall, F1 score, and comparisons against official or manually coded labels [8], [9], [11], [13], [22], [24], [25], [29]. Several studies also show that uncertainty can arise before modeling begins: narratives may be short, vague, ambiguous, inconsistent, or written in specialized sublanguages; human coders may disagree; and official labels themselves may be noisy or incomplete [7], [9], [10], [11], [18], [22], [29]. This means that disagreement between model-generated coding and official database coding cannot automatically be interpreted as model error. In some cases, the narrative may reveal miscoding in the database; in others, the official field may contain information not stated in the narrative; and in still others, both sources may be ambiguous.

This creates the central gap addressed by the present study. The existing literature establishes that crash narratives are useful, that automated methods can extract selected crash attributes, and that LLMs are increasingly relevant to transportation safety analysis. However, it has not yet systematically evaluated the agreement between LLM-generated coding and official database coding across a broad set of official fatal crash attributes under a common prompting protocol, with comparisons against transparent baselines and explicit evaluation of abstention behaviour. Prior studies tend to examine specific crash types, specific severity outcomes, topic patterns, semantic retrieval, or selected reconstruction variables [8], [9], [11], [13], [16], [22], [28]. LLM-focused transportation studies demonstrate promise but often emphasize prediction, review, methodological opportunity, or benchmark hallucination rather than variable-level agreement with official crash database coding [12], [19], [30]. What remains unclear is how agreement with official coding varies across crash attributes and models, how LLM performance compares with transparent non-LLM baselines, and how models differ in coverage, abstention, and agreement among committed outputs under a common prompt.

The present study contributes to this literature by evaluating LLMs as structured crash attribute coders rather than solely as text classifiers or summarization tools. Using a common zero-shot protocol, it compares model-generated classifications with corresponding fields in the official crash database across multiple attributes and examines how agreement varies by model and attribute. The study also evaluates coverage, abstention, and selective agreement to distinguish how often models assign a specific code from how often those committed outputs match official coding. This approach extends prior crash narrative research by moving beyond single-variable classification, benchmarking multiple frontier models against transparent baselines, and evaluating both overall and attribute-specific agreement within a common framework. In doing so, the study provides evidence on the current strengths and limitations of LLM-assisted crash coding while avoiding claims about narrative recoverability, hallucination, or the correctness of either the model output or the official database when the two disagree.

## METHODOLOGY

Six large language models were applied to fatal-crash narratives under a common zero-shot protocol, and their coded output was compared with the corresponding fields of the official structured crash database. The primary endpoint is agreement between model output and official coding. The secondary endpoint is abstention behavior: how often a model declines to assign a code, and how its output agrees with official coding on the subset it does code.

The official database is the comparison source/standard for these comparisons but is not an independently validated one. It is produced by human coders from the full crash report, and we cannot establish, from the available files, whether a given disagreement reflects model error, coding error in the database, residual record-linkage error, or a mismatch between the two coding schemes. Consequently, the design supports statements about agreement and abstention and does not support statements about whether a narrative contained the information in question, whether an abstention was warranted, or whether a committed output was fabricated. Endpoints, results, and conclusions are stated in those terms throughout.

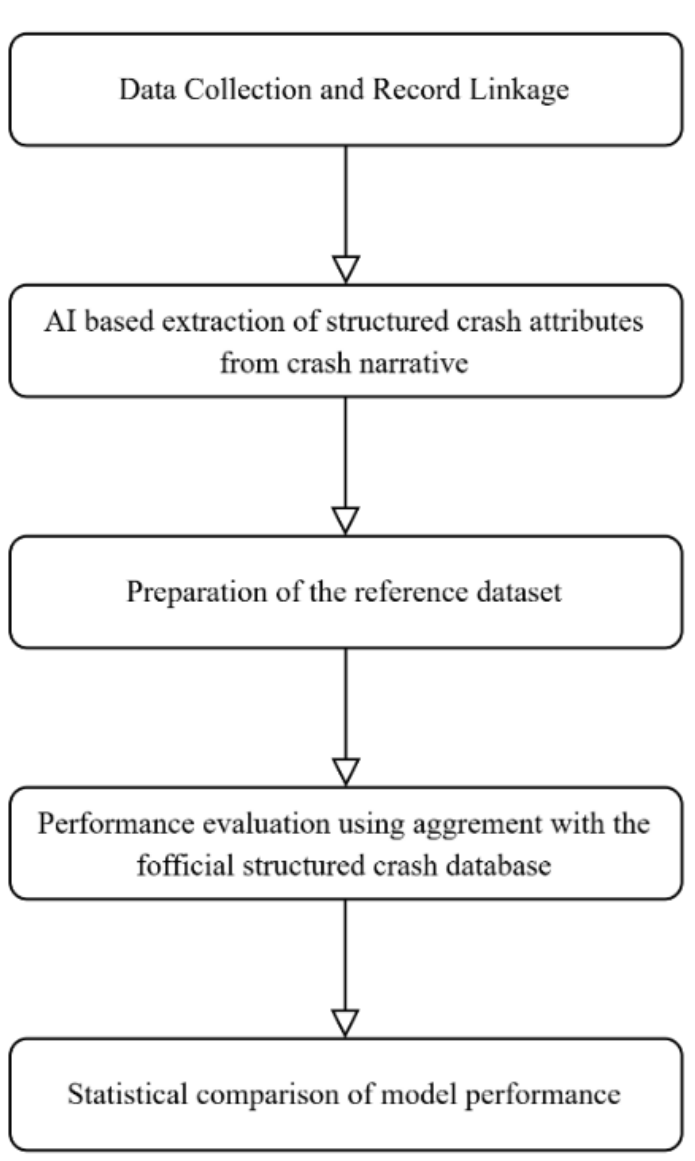

**Figure 1 Methodolgical Workflow**

### Data

Two Arkansas fatal-crash data files covering crashes that occurred between January 1, 2015, and December 31, 2025, were used: 5,587 crash narratives retrieved from the Arkansas Department of Public Safety using a custom Python crawler and 5,889 official structured crash records obtained from the Arkansas Department of Transportation. Because the public narrative database contains only fatal crashes, all Arkansas fatal crashes occurring during the study period were eligible for inclusion. Records without an available narrative or without sufficient date, time, or roadway information for record linkage were excluded. Duplicate records were checked during preprocessing, and none were identified. Both datasets were publicly accessible.

The downloaded data included administrative fields not required for the study, only crash date, reported crash time, roadway location, and narrative text were retained for linkage and model evaluation; the retained narratives contained no personally identifiable information. The narratives were submitted through paid consumer web interfaces for the selected commercial language models rather than through application programming interfaces or enterprise accounts, and no additional provider-specific retention or training controls were applied beyond the default account settings.

The two files are separate extracts but are not independent observations because both originate from the same police crash report, and the officer who wrote the narrative may also have completed the structured fields. Agreement between a model reading the narrative and the structured record may therefore be higher than agreement between genuinely independent sources, while disagreement cannot be attributed definitively to either source without external validation.

Because the files shared no common crash identifier, records were linked using crash date, reported crash time, and street or highway name. Candidate pairs required an exact date match, reported times within ±5 minutes, and roadway names containing common textual elements after punctuation was removed. The time tolerance was used because reported times could differ slightly between the two systems. Candidate pairs were assigned one-to-one in ascending order of absolute time difference, and residual ambiguous cases were independently reviewed by two reviewers using the complete roadway descriptions; both reviewers reached the same linkage decisions. This process produced 4,230 one-to-one pairs, of which 4,194 contained fully numeric codes for all evaluated attributes and formed the final analysis sample, representing 75.1% of narrative records and 71.2% of structured records. Among the matched pairs, 2,883 (68.2%) had identical reported times and 1,347 differed by 1–5 minutes. These counts refer to all 4,230 linked pairs. After the 36 records without fully numeric attribute codes were excluded, the 4,194-record analysis sample contained 2,850 exact-time matches and 1,344 matches differing by 1–5 minutes. Ambiguity was uncommon: 4,208 narratives had exactly one candidate within the linkage window, whereas 86 had two or more candidates. Among unmatched narratives, 1,160 had no structured record within the ±5-minute window on the same date, 107 had no structured record on the same date, 26 had unparseable times, and 64 lost all candidate matches to narratives with smaller time differences. Matched and unmatched narratives had similar mean lengths of 35.1 and 34.4 words, respectively; however, this comparison provides only a limited check on comparability because narrative length may not capture the characteristics associated with linkage success. As a sensitivity analysis, the complete model comparison was repeated using only the 2,850 complete-case crashes with exact reported-time agreement. Agreement for every model changed by less than 0.4 percentage points, and the model ranking remained unchanged. Because no independent manually verified validation sample was created, a separate estimate of linkage accuracy could not be calculated.

**AI-Based Extraction Protocol**

Six models were evaluated: GPT-5.5 (High), GPT-5.5 Instant, Claude Opus 4.8, Claude Sonnet 5, Claude Fable 5, and Gemini 3.1 Pro. The public-facing model names displayed in the provider interfaces are reported because more specific backend version identifiers were not available. All models were accessed through paid consumer web interfaces rather than application programming interfaces or enterprise platforms. No sampling or generation settings were modified; each model was operated using the default settings provided by its web interface. Consequently, parameters such as temperature, top-p, random seed, and provider-side model configuration were neither controlled nor recorded.

The narratives were divided into annual files, with all available narratives from one calendar year uploaded together. Each annual file was processed in a new conversation, resulting in separate model runs for each year from 2015 through 2025. Cross-conversation memory was disabled so that information from one annual batch would not carry into subsequent conversations. Each model received the same annual files, narratives in the same order, identical zero-shot instructions, and no examples or follow-up questions. The prompt directed the model to analyze each narrative independently, rely only on information explicitly stated or strongly supported by the narrative, and return code 999 when an attribute could not be determined.

Each input record contained a unique identifier that was required in the corresponding model output. Following every annual run, automated validation confirmed that the number of output records equaled the number of input records, every input identifier appeared exactly once, no identifiers were missing or duplicated, and all classifications used permitted values from the coding scheme.

Two qualifications apply to this protocol. First, the instruction not to use outside knowledge was a prompt directive rather than a technical constraint. Latent model knowledge could not be disabled, and models may have relied on general crash patterns despite the instruction. Second, the phrase “explicitly stated or strongly supported” established a partly subjective abstention threshold that individual models could interpret differently. Differences among models therefore reflect both their extraction behavior and their interpretation of when the available evidence justified a classification. The study design cannot separate these mechanisms. In addition, because each annual file was processed only once, run-to-run variability was not measured. The use of public model names without backend version identifiers or access dates further limits exact replication because commercial web-based models and their default configurations may change over time. The extracted attributes included: Crash Manner, Light Condition, Intersection Type, Work Zone Relation, Roadway Surface Condition, and Non-Motorist Relation.

**Performance Evaluation**

A consistent metric set was used throughout the Methods, Results, tables, and supplement. For each model and attribute, agreement was reported as the proportion of model outputs matching the official code, but it was not treated as the primary metric because of substantial class imbalance and served as the main descriptive endpoint. Macro-F1 and Cohen's kappa were reported as complementary imbalance-aware measures because both reduce the influence of majority-class prediction. Coverage was defined as the proportion of outputs assigned a specific code rather than 999, and selective agreement measured agreement among those committed outputs. Specific-output disagreement was calculated as coverage × (1 − selective agreement). Official-Unknown agreement measured how often the model returned 999 when the official record was also coded Unknown.

Abstention-related measures were pooled across all 25,164 evaluated attribute values (4,194 crashes × 6 attributes), rather than averaged across attributes, to preserve consistency among coverage, selective agreement, and specific-output disagreement.

**Statistical Analysis**

Each crash yielded one agreement score per model, computed as the proportion of the six attributes for which the model output matched the official code. This composite score served as the outcome for the crash-level comparisons. Mauchly's test assessed sphericity, and the Greenhouse–Geisser correction was applied to the repeated-measures ANOVA. Pairwise differences between models were evaluated with paired t-tests on within-crash differences, adjusted by the Holm procedure and reported as percentage-point differences with 95% confidence intervals. Because every crash contributes an observation to each model, all comparisons treat the data as paired. The Friedman test, with Kendall's W as the effect size, served as a nonparametric check.

The composite score summarizes overall performance but is limited as an inferential outcome: it averages six attributes that differ in class prevalence, category structure, and how often the narrative addresses them, and it takes only seven distinct values. As a confirmatory sensitivity analysis, agreement was also modeled directly. A logistic regression was fitted by generalized estimating equations (GEE) to the binary agreement outcome at the crash-by-model-by-attribute level (150,984 observations), with main effects for model and attribute and their interaction. An exchangeable working correlation structure, with standard errors clustered on crash, accounts for the dependence among the 36 observations that each crash contributes. This specification preserves the model-by-attribute structure that the composite averages away and was used to confirm whether the overall conclusions were consistent when agreement was analyzed without averaging across attributes.

Confidence intervals for the composite scores and attribute-level agreement were obtained by bootstrap resampling of crashes rather than of individual attribute values, so that all six attributes remained clustered within each resampled crash. Percentile intervals were computed from 2,000 replicates for composite scores and 1,000 replicates for attribute-level agreement, under a fixed random seed. Exact Clopper–Pearson intervals were used for the official-Unknown proportions, where the denominators are small. Because the sample contains 4,194 paired crashes, differences of negligible practical magnitude reach statistical significance; absolute percentage-point differences are therefore reported alongside every test.

**Keyword Baseline and Subgroup Definitions**

The keyword baseline applies ordered regular expressions to the upper-cased narrative, returning the first matching code and otherwise a default. For non-motorist relation: bicycle or cyclist terms return Bicyclist, pedestrian terms return Pedestrian, default None. For crash manner: head-on terms return Front-to-Front, sideswipe returns Sideswipe, rear-end terms return Front-to-Rear, angle or T-bone terms return Angle, run-off-road and overturn terms return Single Vehicle, default Single Vehicle. For work-zone relation: work zone or construction terms return Yes, default No. For intersection type: the term intersection returns Four-way, default Not an Intersection. For light condition: dark or night terms return Dark–Not Lighted, daylight returns Daylight, dawn or dusk returns Dusk, default 999. For roadway surface: ice or frost returns Ice/Frost, snow returns Snow, wet or rain returns Wet, dry returns Dry, default 999. The keyword rules and default categories were specified before examining the LLM results and were not revised based on the observed comparative performance. These rules are deliberately crude; their purpose is to establish what a simple transparent system achieves on this task, not to compete as a method.

Subgroup analyses use official database fields for crash configuration, area type, light condition, and surface condition. Because crash manner, light condition, and surface condition are themselves evaluated attributes, subgroup breakdowns defined by them are descriptive locators of disagreement and not independent difficulty measures; they are reported as such. Roadway type is derived from the narrative text by keyword search for interstate, US highway, state highway, county road, and city street terms, with the residual classed as unspecified. This derivation is unvalidated and is reported only as an exploratory stratification.

## RESULTS

### Descriptive Results

The analysis sample comprises 4,194 matched fatal crashes and 25,164 attribute values. Narratives average 35.2 words (median 32) and 57.6% of crashes are rural. Reference class imbalance is severe: 54.5% of crashes are coded single-vehicle, 55.3% daylight, 80.2% non-intersection, 97.9% not work-zone-related, 83.9% dry surface, and 87.1% no non-motorist. The official database codes an attribute Unknown in only 50 of the 25,164 values, concentrated in light condition (19), intersection type (19), and roadway surface condition (12); crash manner, work-zone relation, and non-motorist relation contain no Unknown values.

### Models Against Baselines

**Table 1** places the six models against the three baselines. The result is central to interpreting everything that follows: on raw agreement, the always-majority baseline (0.765) outperforms every model, the best of which reaches 0.385. This is a direct consequence of class imbalance combined with the models' abstention behavior — a system that always predicts the modal class is right whenever the modal class is right, while a model that abstains is scored as disagreeing. On chance-corrected agreement the ordering reverses: the majority baseline scores kappa = 0.000 by construction, whereas the models reach 0.098 to 0.275. Raw agreement and kappa therefore rank these systems in opposite orders, and reporting either alone would misrepresent the comparison.

The keyword baseline is the more demanding comparator. It attains higher raw agreement (0.571) than any model, higher macro-F1 (0.226) than five of six models, and kappa (0.264) below only GPT-5.5 (High) (0.275). A crude regular-expression system is thus competitive with frontier models on this task as measured against the official database. Part of this reflects the keyword system's use of majority-class defaults on three attributes, which raw agreement rewards; but its macro-F1 and kappa are not explained by defaults alone. Any claim that these models add value over conventional text processing for this task is not supported by these results.

**TABLE 1 Models and baselines, averaged over the six attributes**

| System | Agreement | Macro-F1 | Cohen's κ | Coverage |
|---|---|---|---|---|
| Baseline: always-majority | 0.765 | 0.163 | 0.000 | 1.000 |
| Baseline: keyword rules | 0.571 | 0.226 | 0.264 | 0.672 |
| GPT-5.5 (High) | 0.385 | 0.211 | 0.275 | 0.432 |
| GPT-5.5 Instant | 0.378 | 0.188 | 0.266 | 0.428 |
| Claude Sonnet 5 | 0.291 | 0.121 | 0.195 | 0.328 |
| Claude Fable 5 | 0.280 | 0.141 | 0.237 | 0.294 |
| Claude Opus 4.8 | 0.280 | 0.129 | 0.228 | 0.296 |
| Gemini 3.1 Pro | 0.227 | 0.110 | 0.098 | 0.251 |
| Baseline: always-Unknown | 0.002 | 0.001 | 0.000 | 0.000 |

### Agreement by Attribute

Agreement varied more across attributes than across models (**Table 2**). Agreement was highest for non-motorist relation (0.582–0.975), followed by crash manner (0.534–0.788) and intersection type (0.005–0.443). Agreement was substantially lower for work-zone relation (0.002–0.092), roadway surface condition (0.008–0.014), and light condition (0.005–0.006). For work-zone relation, roadway surface condition, and light condition, the low agreement largely occurred alongside very low coverage, indicating that models frequently returned 999 rather than assigning a specific category.

**TABLE 2 Agreement with official coding by attribute and model, with the always-majority rate for comparison**

| Attribute | GPT-5.5 (H) | GPT-5.5 (I) | Sonnet 5 | Opus 4.8 | Fable 5 | Gemini 3.1 | Majority rate |
|---|---|---|---|---|---|---|---|
| Non-motorist relation | 0.975 | 0.972 | 0.926 | 0.960 | 0.973 | 0.582 | 0.871 |
| Crash manner | 0.788 | 0.758 | 0.692 | 0.695 | 0.682 | 0.534 | 0.545 |
| Intersection type | 0.443 | 0.431 | 0.111 | 0.005 | 0.005 | 0.142 | 0.802 |
| Work-zone relation | 0.088 | 0.088 | 0.002 | 0.002 | 0.002 | 0.092 | 0.979 |
| Roadway surface condition | 0.012 | 0.012 | 0.009 | 0.014 | 0.013 | 0.008 | 0.839 |
| Light condition | 0.006 | 0.006 | 0.005 | 0.005 | 0.005 | 0.005 | 0.553 |

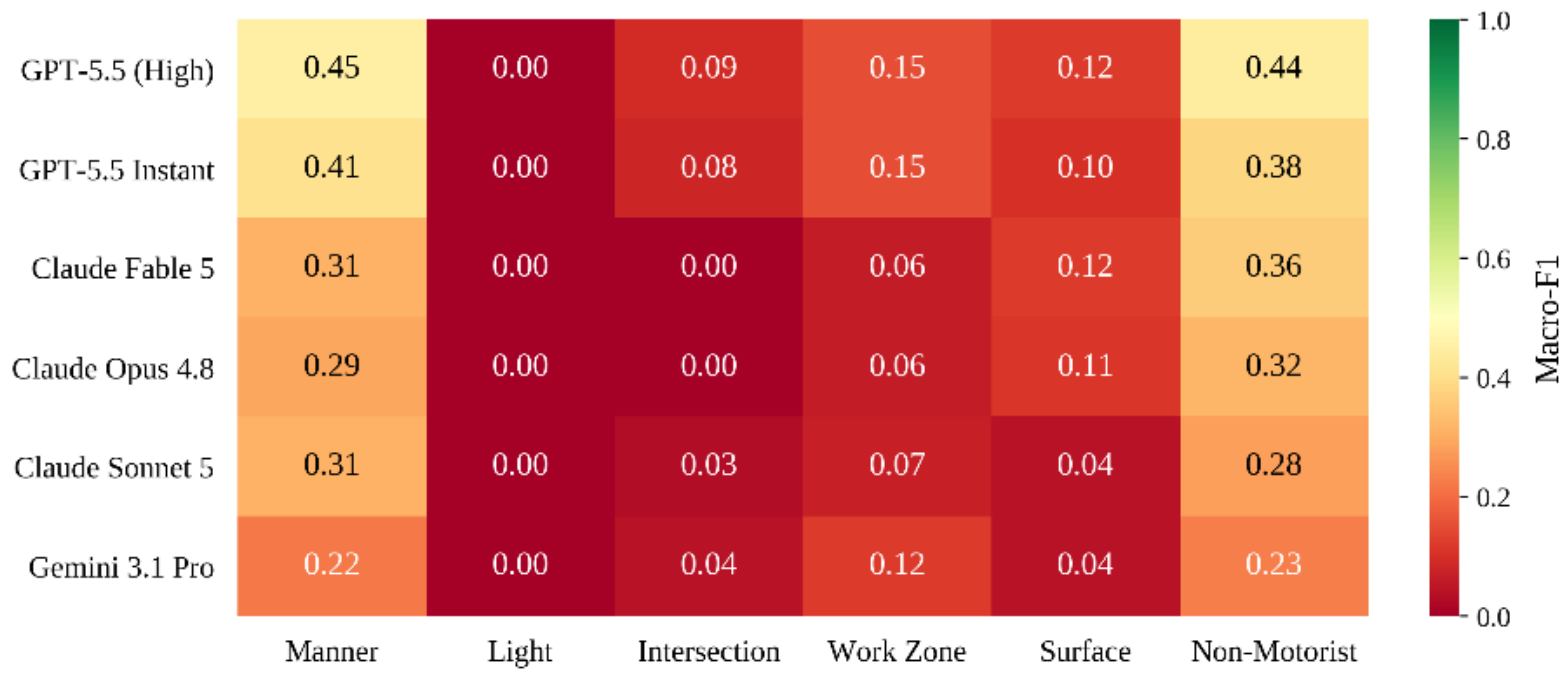

**Figure 2. Macro-F1 by model and attribute**

**Model and Attribute Differences in Agreement**

The statistical analysis examined whether agreement with the official crash database differed across the six models and whether those differences were consistent across the six crash attributes. The repeated-measures analysis confirmed that overall agreement varied significantly among models, $F(2.74, 11{,}482) = 1{,}930.8$, $p < 0.001$, partial $\eta^2 = 0.315$. A nonparametric Friedman test produced the same conclusion, $\chi^2(5) = 6{,}765.5$, $p < 0.001$, Kendall's $W = 0.323$. Pairwise comparisons showed that the two GPT-5.5 models generally achieved higher agreement than the Claude models by 8.7–10.5 percentage points and higher agreement than Gemini 3.1 Pro by 15.1–15.8 percentage points. Differences among the Claude models were small, and Claude Fable 5 and Claude Opus 4.8 did not differ significantly.

Because the overall comparison averages results across attributes with very different coding structures and class distributions, a generalized estimating equations model was also used to evaluate agreement for each crash, model, and attribute combination. The results showed significant differences by model, by crash attribute, and in the interaction between model and attribute (all $p < 0.001$). The interaction indicates that model performance depended on the attribute being coded. For example, models that performed similarly for non-motorist relation or crash manner did not necessarily perform similarly for intersection type, work-zone relation, roadway surface condition, or light condition. Descriptively, the difference between the highest- and lowest-agreement attributes within each model ranged from 0.577 to 0.971. By comparison, the difference between the highest- and lowest-performing models within each attribute ranged from 0.001 to 0.438, indicating greater variation across attributes than across models (**Table 2**).

**Coverage and Committed-Output Agreement**

All six models abstained on more than half of the evaluated attribute values. Coverage ranged from 0.251 for Gemini 3.1 Pro to 0.432 for GPT-5.5 (High), corresponding to 6,309 to 10,881 committed outputs (**Table 3**). Selective agreement on each model's committed subset ranged from 0.879 to 0.945, while specific-output disagreement represented 1.6% to 5.2% of all evaluated values.

Coverage and selective agreement showed a clear tradeoff. Claude Fable 5 and Claude Opus 4.8 had the lowest coverage among the Claude models (0.294 and 0.296) but the highest selective agreement (0.945 and 0.939). GPT-5.5 (High) covered a larger share of values (0.432) but had selective agreement of 0.887. Because each model chose a different subset of cases on which to commit, selective agreement is not a direct like-for-like ranking. The two measures must be considered jointly: higher selective agreement may reflect a more restrictive abstention threshold rather than superior agreement at the same level of coverage.

**TABLE 3 Pooled Coverage and Agreement Among Committed Outputs**

| Model | Coverage | Committed values | Selective agreement | Specific-output disagreement |
|---|---|---|---|---|
| GPT-5.5 (High) | 0.432 | 10,881 | 0.887 | 0.049 |
| GPT-5.5 Instant | 0.428 | 10,775 | 0.879 | 0.052 |
| Claude Sonnet 5 | 0.328 | 8,254 | 0.881 | 0.039 |
| Claude Opus 4.8 | 0.296 | 7,452 | 0.939 | 0.018 |
| Claude Fable 5 | 0.294 | 7,404 | 0.945 | 0.016 |
| Gemini 3.1 Pro | 0.251 | 6,309 | 0.898 | 0.026 |

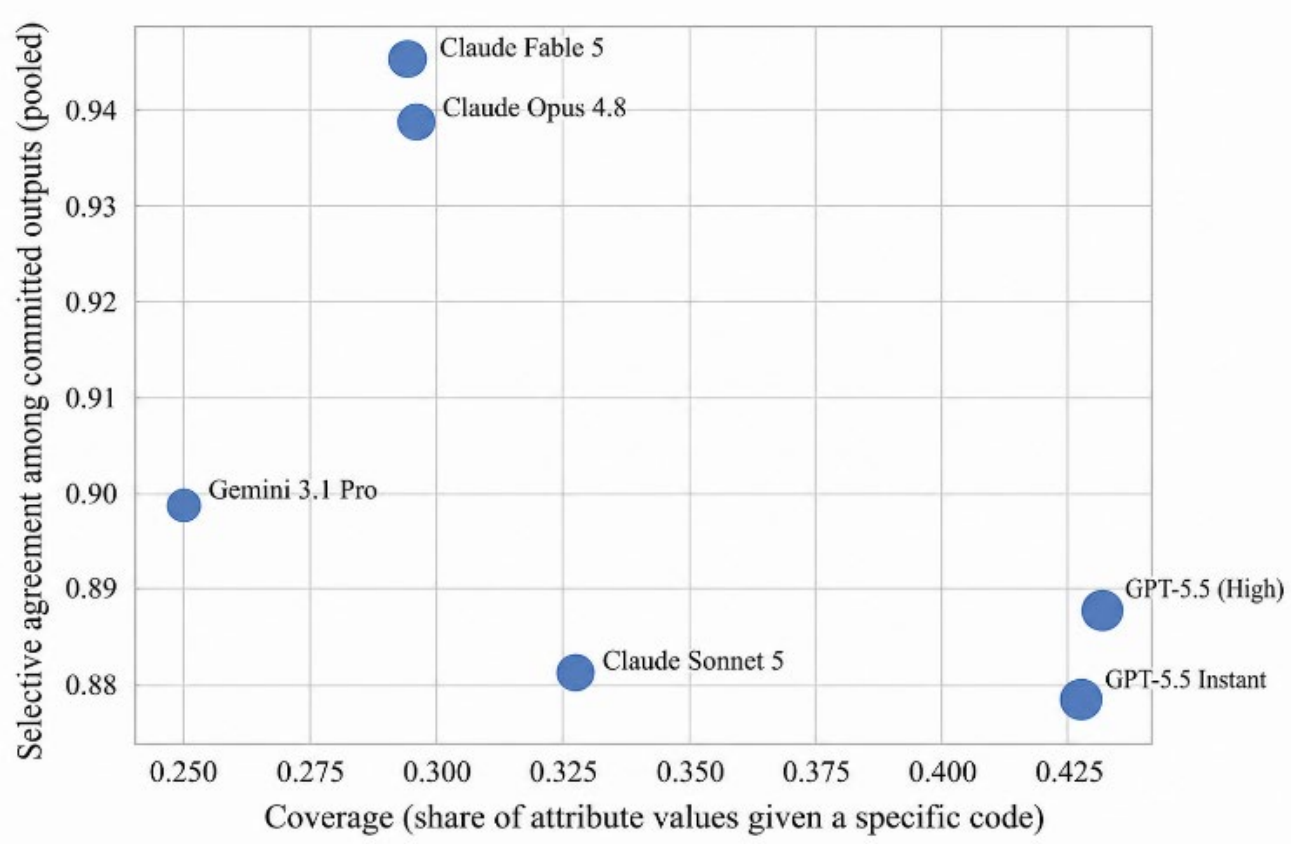

**Figure 3 Coverage against pooled selective agreement; point size is the number of committed outputs.**

Official-Unknown agreement was evaluated by attribute because only 50 official Unknown values were available. All six models returned 999 for all 19 Unknown light-condition records (95% CI 0.82-1.00) and all 12 Unknown roadway-surface records (0.74-1.00). For the 19 Unknown intersection-type records, agreement ranged from 9 of 19 for GPT-5.5 (High) to 19 of 19 for Claude Opus 4.8 and Claude Fable 5. The wide intervals do not support model ranking. Moreover, high agreement on Unknown light and surface values occurred in attributes for which the models abstained on nearly all records, regardless of the official code; it therefore cannot be interpreted as evidence of uncertainty recognition

### Cross-Model Unanimity and Shared Abstention

The six models often produced identical outputs, but the meaning of unanimity differed across attributes (**Table 4 and Figure 4**). For light condition, all models agreed with one another on 99.5% of crashes, and nearly all of that unanimity consisted of shared 999 responses. The same pattern was observed for roadway surface condition (95.9% unanimous; 95.5 percentage points shared abstention), work-zone relation (82.1%; 81.9 points), and intersection type (41.4%; 41.3 points).

Substantive unanimity was concentrated in crash manner and non-motorist relation. All six models assigned the same specific code and agreed with the official record for 43.5% of crash-manner values and 55.4% of non-motorist-relation values. Unanimous specific disagreement was uncommon, reaching 1.7% for crash manner and 0.7% for non-motorist relation. These results show that high cross-model consistency for environmental and work-zone attributes primarily reflected shared abstention rather than agreement on a specific classification.

**TABLE 4 Decomposition of Six-Model Unanimity**

| Attribute | All six identical | All abstain | All agree, specific | All same, different | Models differ |
|---|---|---|---|---|---|
| Light condition | 0.995 | 0.995 | 0.000 | 0.000 | 0.005 |
| Roadway surface condition | 0.959 | 0.955 | 0.004 | 0.000 | 0.041 |
| Work-zone relation | 0.821 | 0.819 | 0.002 | 0.000 | 0.179 |
| Non-motorist relation | 0.562 | 0.000 | 0.554 | 0.007 | 0.439 |
| Crash manner | 0.464 | 0.012 | 0.435 | 0.017 | 0.536 |
| Intersection type | 0.414 | 0.413 | 0.000 | 0.000 | 0.586 |

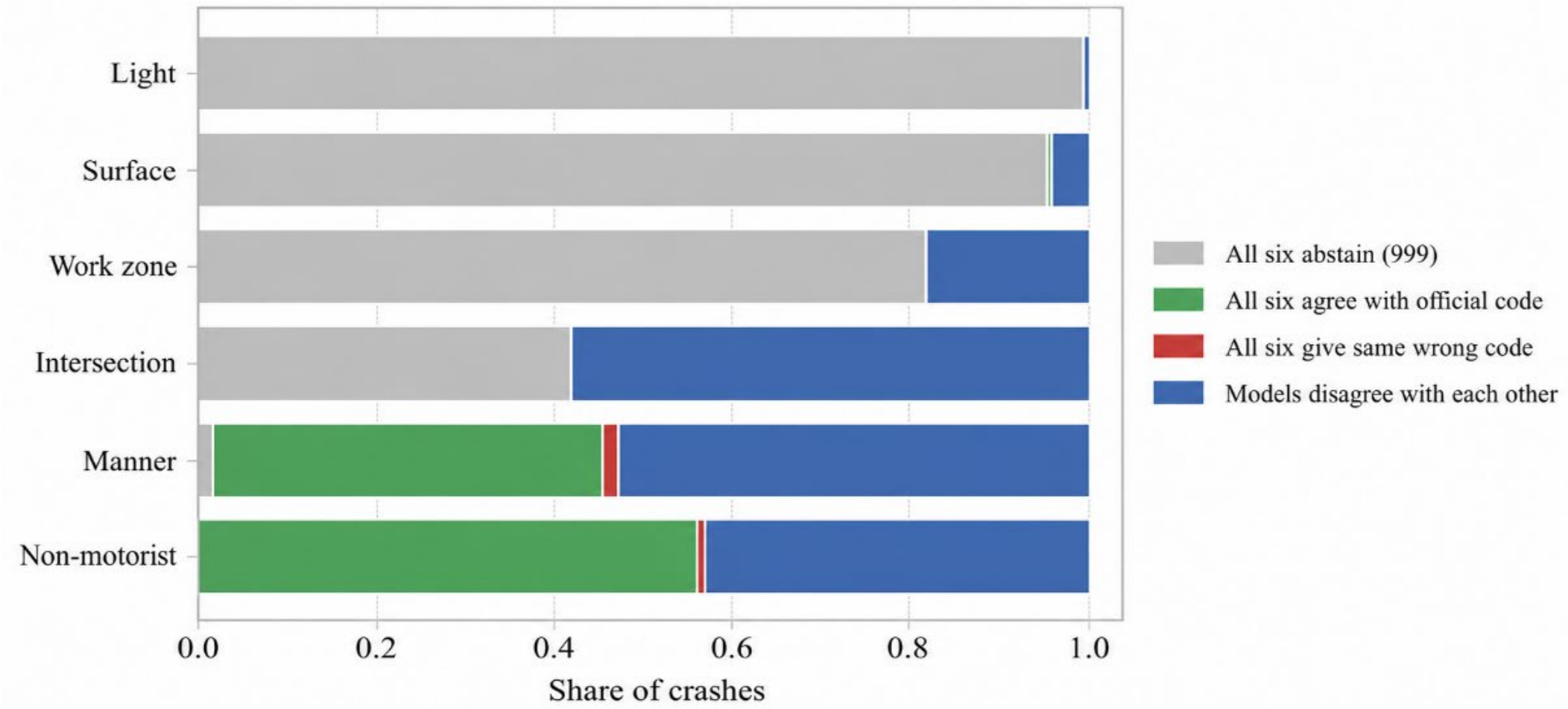

**Figure 4 Unanimity decomposed into shared abstention, shared specific agreement, shared specific disagreement, and mixed model output.**

**Exploratory Patterns**

Exploratory analyses were used to identify where disagreement was concentrated rather than to establish causal explanations. Averaged across the six models, disagreement was lowest for crashes officially coded as single-vehicle (65.5%) and highest for crashes coded Other (79.9%), sideswipe (79.6%), and angle (77.9%). Because these groups were defined using the official crash-manner field, the comparison describes the location of disagreement within that evaluated attribute rather than independent task difficulty.

For GPT-5.5 (High), the highest-agreement model for crash manner, frequent confusion patterns included single-vehicle crashes coded as Other (171 crashes; 7.5% of officially coded single-vehicle crashes), angle crashes coded as Front-to-Front (119; 18.0% of angle crashes), and Front-to-Front crashes coded as angle (62; 9.2%). The reciprocal angle and Front-to-Front pattern is consistent with narratives that do not fully describe vehicle approach geometry, although independent narrative review would be required to confirm that explanation.

Narrative length was formally evaluated in a GEE model that adjusted for model and attribute. A one-standard-deviation increase in narrative length was associated with lower odds of agreement (odds ratio 0.839, 95% CI 0.797-0.883, $p < 0.001$). The association was modest may reflect unmeasured differences in crash complexity or other characteristics associated with longer narratives. Descriptive disagreement varied from 67.7% to 71.6% across keyword-derived roadway types and was 68.4% in rural areas and 70.6% in urban areas; these subgroup differences were not formally tested and are reported in the supplement.

**DISCUSSION**

The findings address the three study questions from complementary perspectives. Among the evaluated large language models, the GPT-5.5 variants achieved the highest agreement with official coding; however, comparisons with simple baselines changed the interpretation of this ranking. The always-majority baseline produced the highest raw agreement because the official data were highly imbalanced, while a transparent keyword-rule system achieved chance-corrected agreement comparable with the best-performing large language model. Agreement also varied more across crash attributes than across models, and the significant model-by-attribute interaction showed that no single model ranking applied consistently to all coded fields. In addition, the models differed substantially in how often they assigned specific codes rather than abstaining. Shared abstention accounted for most cross-model unanimity for light condition, roadway surface condition, work-zone relation, and intersection type.

### Interpretation of Model and Attribute Differences

The model comparison should not be reduced to a single overall score. GPT-5.5 (High) and GPT-5.5 Instant consistently led the LLM group on the composite outcome, but their advantage varied by attribute. The largest separation occurred for intersection type, whereas model differences were comparatively small for roadway surface and light condition because all models returned 999 on nearly all records. The significant interaction formalizes this practical point: transportation agencies cannot assume that a model selected on overall performance will be the best option for every crash-data field.

The attribute pattern is equally important. Non-motorist relation and crash manner showed substantially higher agreement with official coding than the environmental and work-zone attributes. This may reflect differences in how crash narratives are written: road-user involvement and collision configuration are often central to the event description, while lighting, surface, and work-zone status may be recorded primarily in structured fields. However, the study did not independently label narrative evidence, so this explanation remains a hypothesis rather than a demonstrated finding. The results establish where agreement is high or low, not why the pattern occurs.

### Abstention and the Meaning of Selective Agreement

The coverage results show that the evaluated LLMs behaved as selective coders rather than complete database-population tools. Even the highest-coverage model assigned a specific code for only 43.2% of values. Models with lower coverage often achieved higher agreement on the cases they chose to code, illustrating a familiar coverage-performance tradeoff. For transportation applications, this distinction matters because a high selective-agreement value can coexist with limited operational usefulness when most records are left unresolved.

The unanimity analysis further shows why model consensus must be interpreted carefully. Near-unanimous output for light and surface condition did not indicate that the models consistently identified the same roadway condition; it indicated that they consistently declined to code the field. In contrast, unanimity for non-motorist relation and crash manner more often represented the same specific classification and agreement with the official record. A useful evaluation framework therefore needs to distinguish shared classification from shared abstention rather than treating all agreement among models as equivalent.

## IMPLICATIONS FOR TRANSPORTATION AGENCIES

The present results do not support direct deployment of these models to populate structured crash databases. The study did not include a human narrative-coding benchmark, a prospective agency workflow, time or cost measures, or an acceptable-error threshold. Without those elements, agreement with the official database cannot be translated into operational accuracy, staff savings, or quality-control value.

The findings nevertheless identify a more focused evaluation pathway. Non-motorist relation and crash manner are reasonable candidates for further human-adjudicated evaluation or a limited pilot because several models showed relatively high agreement and substantial specific-output unanimity for those attributes. Such a pilot should measure the number and quality of discrepancies flagged for review, the false-positive burden placed on staff, and whether model assistance changes coding time or consistency. Environmental and work-zone attributes would require a different strategy because the current prompt produced little usable coverage.

The baseline results also have direct practical importance. A simple rule-based system was competitive with the LLMs on several summary measures. Transportation agencies considering automated narrative coding should therefore compare LLMs with transparent conventional methods under the same data, attribute definitions, and workflow requirements. A more complex model is justified only if it provides measurable gains in accuracy, coverage, maintainability, or staff effort.

## LIMITATIONS

The primary limitation is the absence of an independently adjudicated reference. The official structured database is the only comparator, and disagreement may reflect model output, database coding, linkage error, or differences between the information recorded in narrative and structured fields. Because the narrative and structured fields arise from the same crash-reporting process, they are not independent sources, and agreement may be higher than would be observed against an external reference.

Record linkage and sample selection introduce additional uncertainty. Approximately 25% of narratives were unmatched. The exact-time sensitivity analysis produced nearly identical model rankings, but matched and unmatched records were compared only on narrative length, and the linkage procedure has not yet been validated against a manually reviewed sample. The findings also depend on one prompt, one run per model, and model versions available at a single point in time. Proprietary systems may change, sampling settings and system instructions differ across providers, and repeatability was not measured.

A human-adjudicated reference set would be the most informative next study for extending the agreement benchmark. A feasible study could use a stratified sample of approximately 400 to 600 crashes including cases of shared abstention, shared disagreement, and mixed-model output. Blinded reviewers could classify the level of narrative support for each attribute, with disagreements resolved by a third reviewer and inter-rater reliability reported by attribute.

## CONCLUSION

Six LLMs were evaluated on 4,194 linked Arkansas fatal-crash narratives and six structured attributes. Overall agreement with official coding ranged from 0.227 to 0.385 across the LLMs, with GPT-5.5 (High) and GPT-5.5 Instant leading the model group. Agreement varied much more across attributes: non-motorist relation and crash manner showed the highest agreement, while light condition, roadway surface condition, and work-zone relation showed near-zero agreement because models frequently returned 999, which was scored as disagreement whenever the official record contained a specific code. Coverage ranged from 25.1% to 43.2%, and model rankings changed across attributes.

The baseline comparisons limit the conclusions that can be drawn from the model ranking. An always-majority system achieved the highest raw agreement, and a simple keyword extractor achieved chance-corrected agreement close to the best LLM. The study therefore provides an agreement and abstention benchmark rather than evidence that current LLMs are ready to code crash databases from narratives. Its main contribution is to identify which attributes and disagreement patterns require independent human adjudication before transportation agencies can evaluate accuracy, workflow value, or deployment readiness.

## ACKNOWLEDGMENTS

OpenAI ChatGPT 5.5 was used to assist with grammar and language editing, improve the clarity of the manuscript, and support the development and debugging of computer code used in the analyses. The study design, data preparation, analyses, interpretation of the results, and all final decisions regarding the manuscript content were performed by the authors, who reviewed and verified all AI-assisted outputs.

## DISCLAIMER

The contents of this report reflect the views of the authors, who are responsible for the facts and the accuracy of the data presented herein. The contents do not necessarily reflect the official views or policies of ARDOT, which assumes no liability for the contents or use thereof. This report does not constitute a standard, specification, or regulation. Comments contained in this report related to specific testing equipment and materials should not be considered an endorsement of any commercial product or service; no such endorsement is intended or implied.

## AUTHOR CONTRIBUTIONS

The authors confirm contribution to the paper as follows: study conception and design: Sudhir Bharati; data collection: Sudip Bharati, R. KC Khatri; analysis and interpretation of results: R. KC Khatri, Sudhir Bharati, Sudip

Bharati; draft manuscript preparation: Sudhir Bharati, R. KC Khatri, Sudip Bharati. All authors reviewed the results and approved the final version of the manuscript.

## DECLARATION OF CONFLICTING INTERESTS

The authors declared no potential conflicts of interest with respect to the research, authorship, and/or publication of this article.

**FUNDING**

The authors disclose no financial support for the research, authorship, and/or publication of this article.